\documentclass{article}
\usepackage{nips13submit_e,times}
\usepackage[latin9]{inputenc}
\usepackage{color}
\usepackage{graphicx, caption, subcaption, paralist}
\usepackage{amsmath,amssymb} 
\usepackage{xspace}
\usepackage{booktabs}
\usepackage{tabularx}
\usepackage{multirow} 

\usepackage{times}\usepackage{epsfig}

\usepackage[unicode=true,pdfusetitle,
 bookmarks=true,bookmarksnumbered=false,bookmarksopen=false,
 breaklinks=false,pdfborder={0 0 1},backref=section,colorlinks=true]
 {hyperref}

\makeatletter

\DeclareRobustCommand\onedot{\futurelet\@let@token\@onedot}
\def\@onedot{\ifx\@let@token.\else.\null\fi\xspace}

\def\ie{\emph{i.e}\onedot}

\def\etal{\emph{et al}\onedot}
\makeatother

\iftrue
\newcommand{\todo}[1]{{\textcolor{red}{[[TODO: #1]]}}}
\newcommand{\outline}[1]{{\textcolor{blue}{[[#1]]}}}
\newcommand{\commenttext}[1]{\textcolor{red}{[[#1]]}}
\newcommand{\commentfoot}[1]{\footnote{\textcolor{red}{#1}}}
\newcommand{\commentselfoot}[2]{{\textcolor{blue}{#1}}\comment{#2}}
\newcommand{\commentselrep}[2] {{\textcolor{blue}{#1}} {\textcolor{green}{[[\textit{#2}]]}}}
\else
\newcommand{\todo}[1]{}
\newcommand{\outline}[1]{}
\newcommand{\commenttext}[1]{}
\newcommand{\commentfoot}[1]{}
\newcommand{\commentselfoot}[2]{}
\newcommand{\commentselrep}[2]{}
\fi

\makeatother

\title{Learned versus Hand-Designed Feature Representations for 3d Agglomeration}

\author{
John A. Bogovic, Gary B. Huang \& Viren Jain \\
Janelia Farm Research Campus \\ 
Howard Hughes Medical Institute \\
19700 Helix Drive, Ashburn, VA, USA \\
\texttt{\{bogovicj, huangg, jainv\}@janelia.hhmi.org} 
}

\nipsfinalcopy 

\begin{document}
\maketitle

\begin{abstract}
For image recognition and labeling tasks, recent results suggest that
machine learning methods that rely on manually specified feature
representations may be outperformed by methods that automatically
derive feature representations based on the data. Yet for problems
that involve analysis of 3d objects, such as mesh segmentation, shape
retrieval, or neuron fragment agglomeration, there remains a strong
reliance on hand-designed feature descriptors.  
In this paper, we evaluate a large set of hand-designed 3d feature
descriptors alongside features learned from the raw data using both end-to-end and unsupervised learning techniques, in the context
of agglomeration of 3d neuron fragments. By combining unsupervised learning techniques with a novel dynamic pooling scheme, we show how pure learning-based methods are for the first time competitive with hand-designed 3d shape descriptors. We investigate data augmentation strategies for dramatically increasing the size of the training set, and show how combining both learned and hand-designed features leads to the highest accuracy. 

\end{abstract}

\section{Introduction}

A core issue underlying any machine learning approach is the choice
of feature representation. Traditionally, features have been hand-designed
according to domain knowledge and experience (for example, Gabor filters
for image analysis or cepstral coefficients for automatic speech recognition).
Recently, it has become more common to attempt to learn features based
on supervised or unsupervised learning methods~\cite{LeCun:1998,Hinton:2006,bengio2009learning,coates2012emergence,coates2011analysis,Ranzato:2007,farabet2012scene}.
These automatically derived feature representations have the advantage
of not requiring domain expertise and potentially yielding a much
larger set of features for a classifier. Perhaps most importantly,
however, automatic methods may discover features that are more finely
tuned for the particular problem being solved and thus lead to improved
accuracy.

For many problems that involve analysis of 3d objects there remains
a strong reliance on hand-designed feature descriptors even when machine
learning is used in conjunction with such descriptors. For example,
the field of 3d shape retrieval has a substantial history of benchmarking
hand-designed shape descriptors~\cite{li2012shrec,tangelder2008survey}.
Mesh segmentation has recently been addressed using a conditional
random field with energy terms based on a hand-curated set of shape-based
features~\cite{kalogerakis2010learning}. Supervoxel agglomeration
for connectomic reconstruction of neurons has, thus far, also been
largely dependent on manually specified feature representations~\cite{Andres:2008,jain2011learning}. 

\begin{figure}[t]
  \begin{center}
	 \begin{tabular}{ccc}
		\includegraphics[height=1.3in]{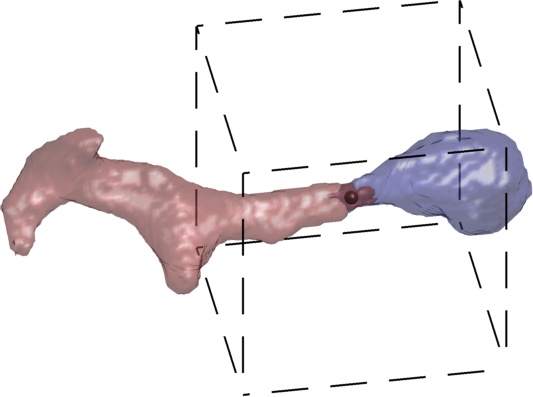} &
		
&
		\includegraphics[height=1.3in]{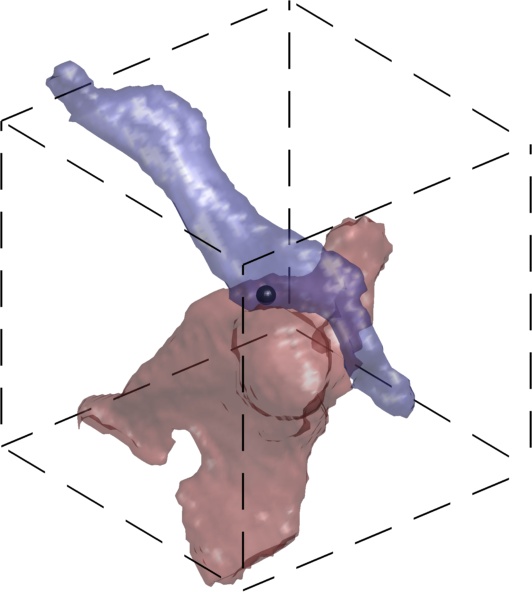} \\ 
		(a) Positive Example &  & (b) Negative Example \\
	 \end{tabular}
  \end{center}
  \caption{\small Renderings of canonical positive and negative edge examples from the training set. We denote a pair of supervoxels which are subject to binary classification as a single \emph{edge} in the overall agglomeration task \cite{jain2011learning}. The small black sphere indicates the decision point around which most computations are centered, and the dotted box indicates a cube with a length of $30$ pixels in each dimension.}
  \label{fig:exampleRenderings}
\end{figure}

Designing features for representing specific kinds of 3d objects is arguably
more intuitive as compared to hand-designing representations for more low-level
data (such as raw image patches). For example, describing a neuron fragment in
terms of quantities such as curvature, volume, and orientation seems natural. On
the other hand, it is less clear this intuitive appeal is a good justification
for such a feature representation in a specific task such as neuron fragment
agglomeration (Figure~\ref{fig:exampleRenderings} and supplementary
Figure~\ref{fig:extra_edge_examples}).

The primary contributions of our work are:

\begin{compactenum}

	\item A large set of diverse hand-designed 3d shape descriptors that dramatically improve performance over simple baseline features used in prior work. We evaluate each feature individually, evaluate an ensemble set of all hand-designed features, and compare the computational cost of the features.
	\item An unsupervised learning approach for deriving 3d feature descriptors that, when combined with a novel dynamic pooling scheme, yields  performance comparable to an ensemble set of all hand-designed features. To our knowledge, this is the first time purely learned features have been shown to provide competitive performance on a task involving analysis or classification of 3d shapes.
	\item An end-to-end supervised learning approach for deriving 3d feature descriptors. We introduce data augmentation strategies that dramatically expand the size of the training set and thus improve generalization performance of the end-to-end feature learning scheme.

\end{compactenum}

\section{Agglomeration of 3d Neuron Fragments}
\label{sec:agglomeration}
We focus on the application domain of segmentation of large-scale
electron microscopy data for the purposes of `connectomic'
reconstruction of nervous system structure. Mapping neural circuit
connectivity at the resolution of individual synapses is an important
goal for neurobiology, which requires nanometer resolution imaging of
tissue over large fields of view
\cite{Helmstaedter:2008}. Interpreting the resulting tera- or
peta-voxel sized datasets currently involves substantial human effort,
and thus increased or complete automation through highly accurate
computational reconstruction would be ideal \cite{jain2010machines}.

Automated pipelines for segmentation of both natural and non-natural
images have converged on a broadly similar set of steps: boundary
prediction, oversegmentation, and agglomeration of segments~\cite{Fowlkes:2003,farabet2012scene,jain2011learning,Andres:2008,chklovskii2010semi}. In this section we describe the source of the raw data, the creation of 3d segments, and the machine learning problem of fragment agglomeration.

\textbf{Electron microscopy images}:
\label{sec:EM}
Tissue from a \emph{drosophila melanogaster} brain was imaged using
focused ion-beam scanning electron microscopy (FIB-SEM \cite{knott2008serial})
at a resolution of $8\times8\times8$ nm. The tissue was prepared
using high-pressure freeze substitution and stained with heavy metals
for contrast during electron microscopy. As compared to traditional
electron microscopy methods such as serial-section transmission electron
microscopy (ssTEM), FIB-SEM provides the ability to image tissue at
very high resolution in all three spatial dimensions. Isotropic resolution
at the sub-$10$nm scale is particularly advantageous in drosophila
due to the small neurite size that is typical throughout the neuropil.

\textbf{Boundary prediction:}
\label{sec:boundary_prediction}
We trained a deep and wide multiscale recursive (DAWMR) network \cite{huang2013deep} to
generate affinity graphs from the electron microscopy data. Affinity
graphs are similar to pixel-wise boundary prediction maps, except
that they encode connectivity relationships between neighboring pixels
(in our case, $6$-connectivity due to the 3d image space) \cite{Turaga:2010uq}.
We supplied the DAWMR network with $120$ megavoxels of hand-segmented image
data for training (with rotation and \emph{x-y} reflection augmentations
further increasing the total amount of data seen during training).  The network
uses a total field of view  of $50\times50\times50$ pixels in the prediction of
any single affinity edge. 

The ground truth affinity graphs are binary representations where
$0$ represents the case where two pixels are disconnected (belong
to different objects, or are both part of `outside' space unassigned
to any object), and $1$ represents the case where two pixels are
part of the same object. The DAWMR networks, trained on this ground
truth, generate analog $[0,1]$-valued affinity graphs.

\textbf{Oversegmentation:}
\label{sec:over_segmentation}
The DAWMR-generated affinity graph is thresholded at a value of $0.9$
and objects are `grown' by a seeded watershed procedure to
an affinity value of $0.8$. The affinity graph is then re-segmented
at $0.8$, new objects are added into the overall segmentation, and
all objects are grown to a threshold of $0.7$. This procedure is repeated
for thresholds $0.6$ and $0.5$. A distance-transform based object-breaking
watershed procedure is then applied that slightly reduced the rate
of undersegmentation in large objects. Finally, all objects are grown
to a threshold of $0.2$.

\textbf{Training and test sets:}
\label{sec:data_sets}
Two separate $200$ megavoxel volumes were processed by the DAWMR
network and oversegmented according to the procedure described above.
Neither volume contained data used to train the boundary predictor.
Pairs of segments within $1$ pixel of each other (we refer to these
identified segment-pairs as \emph{edges}) were labeled by humans as to whether
the segments belong to the same or different neuron.  One of the two volumes was
randomly chosen to be the training set ($14,522$ edges: $7968$ positive and
$6584$ negative), and the other volume serves as a test set ($14,829$ edges:
$8342$ positive and $6487$ negative).  Figure~\ref{fig:exampleRenderings} shows
examples of both positive and negative segment-pairs. 

\textbf{Learning binary agglomeration decisions:}
\label{sec:learning}
Superpixel agglomeration has recently been posed as a machine learning
problem; some methods attempt to optimize classifier performance over
a \emph{sequence} of predictions that reflect, for example, variable ordering
of agglomeration decisions based on classifier
confidence~\cite{Nunez-Iglesias2013,jain2011learning}. In this work, we simply train a classifier on a one-step cost function that reflects the ground truth binary edge assignments. This is designed to simplify the interpretation of feature contributions and ease the computational burden of the many classification experiments we perform. We learn binary agglomeration
decisions using a dropout multilayer perceptron (MLP)~\cite{hinton2012improving}, and for comparison provide certain results using a decision-stump boosting classifier~\cite{Freund1999}.  

\section{Hand-Designed Features} \label{sec:hand_designed_features}

In this section we describe the proposed hand-designed features and
evaluate the performance of each feature by measuring its accuracy
on the agglomeration classification problem.

The features for a given pair of segments are computed from a
fixed-radius subvolume centered around a `decision point' between
the two segments (Figure~\ref{fig:exampleRenderings}).  The subvolume consists of the raw image values as well as
the affinity graph produced by the DAWMR network.  For simplicity,
we often collapse the affinity graph by averaging over the three
edge directions, which we refer to as the `boundary map.'

The decision point is defined as the midpoint of the shortest line
segment that touches both segments. The motivation for this scheme lies
in the intuition (based on observing human classification strategies)
that the relevant image and object information required to decide
whether two segments should be merged is concentrated near the
interface between the two segments. 

\subsection{Feature Descriptions}
\textbf{Boundary map statistics:}\label{sec:baseline} after identifying a
set of pixels that constitute the interface between two segments, we compute a
number of statistics of the boundary map values over these pixels: mean, median,
moments (variance, skewness, kurtosis), quartiles, length, minimum value, and
maximum value.  We also compute these statistics from the first and second
derivative of the boundary map.  This follows many previous approaches that
identify some type of interface between segments, measure statistics at boundary
map locations along this interface, and use these statistics as features to
train an agglomeration
classifier~\cite{Andres:2008,farabet2012scene,jain2011learning}.  The boundary map is obtained by averaging the edges of the affinity
graph.  As noted in
Table~\ref{tab:hand_selected_results}, we consider the ensemble of these
statistics (experiment 6) as a baseline feature set.

\textbf{Size:} the volumes of both segments, and their log value.

\textbf{Proximity:} a scalar giving the shortest distance from a voxel
assigned to one segment to a voxel assigned to the other segment.

\textbf{Growth:}\label{sec:grow_features} segments are isolated within the
component mask and are grown via a seeded watershed transform until they share a
catchment basin. The affinity graph value at which this occurs yields the first
growth feature.  The second growth feature is given by the distance from the
decision point to the location at which the catchment basins merge.

\textbf{Rays:} lines are propagated from the centroid of a segment until they
terminate \cite{Smith2009}. The features describe the average distance these
rays travel before termination under one of two conditions: the affinity graph
value falls below a specified threshold, or the ray exits a mask defined by the
union of the two segments.  We seed rays from both segments and use five
choices of affinity graph threshold. Our experiments used $42$ rays uniformly
distributed over the sphere.

\begin{figure}[t]
  \begin{center}
   \begin{tabular}{ccc}
	\includegraphics[width=0.4\textwidth]{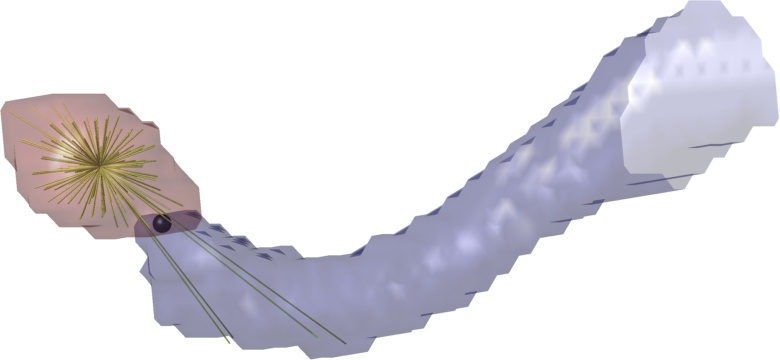} & 
	 \includegraphics[width=0.2\textwidth]{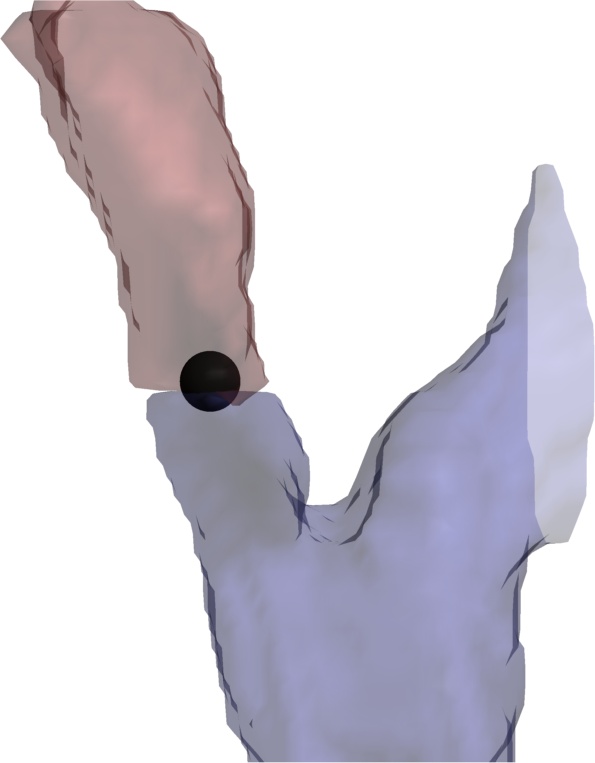} & 
	 \includegraphics[width=0.2\textwidth]{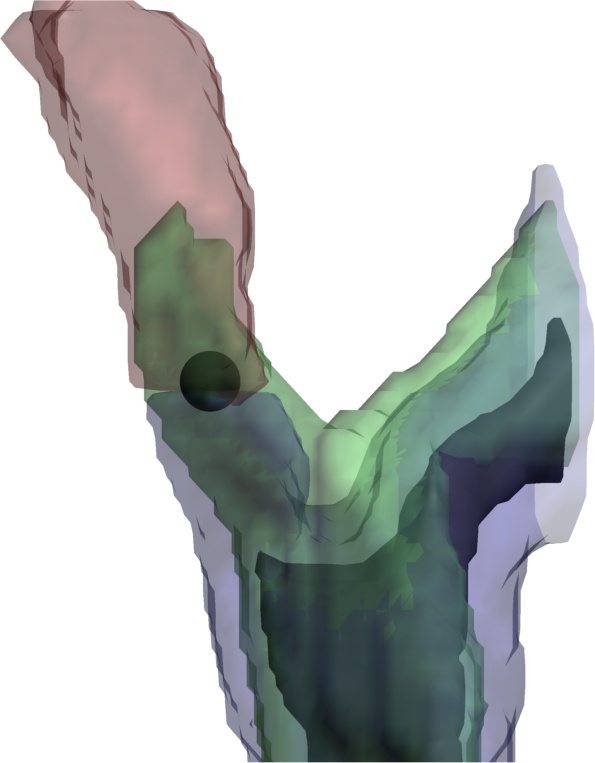} \\
   (a) Ray Features & (b) Level Set Initialization & (c) Level Set Evolution  \\  
  \end{tabular}
  \end{center}
  \caption{\small Demonstration of ray and level set features on positive edge examples. Rays originate in the red segment and penetrate the blue segment. The surfaces in (b) show the initialization state of the level set, and the multiple green surfaces in (c) show the results after various amounts of level set evolution. }
  \label{fig:angleDemo}
\end{figure}

Another type of ray feature describes the average distance rays travel through
one segment when seeded from the other segment.
Figure~\ref{fig:angleDemo}(a) shows an example of the rays used for this feature. 


\textbf{SIFT:} scale-invariant feature transform (SIFT) descriptors are
computed that summarize the image gradient magnitudes and orientations near the
decision point \cite{Lowe2004}. We cluster the descriptors using k-means with
$50$ clusters and represent each descriptor as a feature vector based on a soft
vector quantization encoding. SIFT features are computed using both the image
data and the affinity graph.

\textbf{Angles:}\label{sec:angle_features} we compute two vectors, $v_{o1}$
and $v_{o2}$, giving the orientation of each of the segments and a third vector,
$v_c$, that points from the center of mass of one segment to the center of mass
of the other.  The orientation of each segment is computed from a smooth vector
field determined by the largest eigenvector of a windowed segments'
second-moment matrix (see the Appendix of \cite{jain2011learning} for details).
Features include the length of $v_c$, and the angles formed by $v_{o1}$ and
$v_{o2}$ with $v_c$.  This procedure is repeated with downsampled object masks,
and objects grown using the affinity graph watershed transform (as for growth
features) with $9$ choices of threshold, yielding $33$ angle features in total.
Intuitively, we expect that two segments should be merged if the orientation of
one segment is parallel with the vector pointing to the other segment. 

\textbf{Level sets:} a segment is eroded to produce a contour that
initializes a level set~\cite{Malladi1995}. It is then evolved under a speed function that should,
ideally, result in the deformed segment moving towards and into the other
segment \emph{if} those segments belong together. The speed function determining
the evolution consists of orientation and gradient vector flow fields.


The orientation vector field is computed from the primary eigenvalue of the
second-image-moment matrix. This field serves to move the initial contour from
one segment to the other and provides evidence for positive examples.  The
eigenvalues of the moment matrix describe how tubular, flat, or spherical each
segment is. Therefore, we also compute the mean and standard deviation of the three
eigenvalues yielding $6$ orientation features.

A gradient vector flow (GVF) field \cite{Xu1998} is computed from the boundary 
prediction map in a manner similar to \cite{Yang2013}. This field can prevent
the contour from crossing the boundary between segments and serves as evidence
for negative examples. We compute the mean and standard deviation of the curl
and divergence of the gradient vector field over the interface between
segments, yielding $4$ GVF features.

The level set overlap feature is the number of pixels belonging to both the
level set result and the other segment.  This process is repeated in reverse
(starting the evolution from the other segment).  We use these two overlap
quantities, along with the mean, minimum, maximum and absolute difference
between the two results, to yield $6$ overlap features in total. 

\textbf{Shape diameter function:} the local width of each segment,
represented via statistics on the shape diameter function as defined in
\cite{Shapira2008}. The shape diameter function has been widely used for 3d
mesh analysis and segmentation. We include both moments ($8$ features) and
quantile-based statistics ($10$ features).

\textbf{Shape context:} the local shape of each segment using a 3d implementation 
of \cite{Belongie2002}.  In particular, we consider the shape to be the set of
all points inside the window and on the boundary of either segment.  Shape
context is computed using the window's central point as a reference, and a
histogram with $5$ radial, $12$ polar angle and $12$ azimuth angle bins.
We cluster these quantities using k-means ($20$ clusters) then represent the feature using soft vector quantization. 

\begin{table*}[ht]
\setlength\tabcolsep{3pt}
\centering
\begin{tabular}{ c r l c c c c r r }
\toprule
 & & & \multicolumn{2}{c}{Training set} & \multicolumn{2}{c}{Testing set} & & \\
\cmidrule(r){4-5} \cmidrule(r){6-7} \cmidrule(r){8-9} 
 & Exp.& Feature Set Description & ACC(\%) & AUC(\%) & ACC(\%) & AUC(\%) & Dim. & Cost\\ 
\midrule 
\multirow{8}{*}{ \rotatebox{90}{Boundary Map (bm)}}
 &1& bm mean, median, interface len. & 83.64 & 92.44 & 80.92 & 90.46 & 6 & 5.7 \\ 
 &2& exp 1 + bm moments & 85.27 & 93.39 & 82.68 & 91.64 & 9 & 5.7 \\ 
 &3& exp 1 + bm quantiles & 84.54 & 93.11 & 82.04 & 91.33 & 8 & 5.7 \\ 
 &4& exp 1 + bm quantiles, min/max & 85.03 & 93.36 & 82.69 & 91.63 & 10 & 7.9 \\ 
 &5& exp 1 + bm deriv. mean, median & 90.31 & 96.71 & 88.64 & 95.73 & 14 & 7.9 \\ 
 &6& exp 1 + all bm deriv. stats & 91.85 & 97.61 & 89.11 & 96.05 & 42 & 14.0 \\ 
 &7& baseline ($\bigcup$ exp 1:6) & 92.30 & 97.85 & 89.41 & 96.05 & 49 & 14.0 \\ 
 &8& baseline ($\bigcup$ exp 1:6) boosting & 92.17 & 97.88 & 88.56 & 95.36 & 49 & - \\ 
 \midrule
 \multirow{12}{*}{ \rotatebox{90}{Object}}
 &9& exp 7 + growth & 92.55 & 98.09 & 89.67 & 96.32 & 51 & 1.0 \\ 
 &10& exp 7 + proximity & 92.18 & 97.85 & 89.09 & 96.02 & 50 & 489.1 \\ 
 &11& exp 7 + angles & 95.74 & 99.25 & 89.65 & 96.27 & 82 & 13.0 \\ 
 &12& exp 7 + size & 93.31 & 98.43 & 90.28 & 96.61 & 53 & 1.9 \\ 
 &13& exp 7 + rays & 94.36 & 98.92 & 90.06 & 96.52 & 91 & 44.2 \\ 
 &14& exp 7 + shape diam. quantiles & 93.52 & 98.56 & 89.82 & 96.46 & 59 & 402.5 \\ 
 &15& exp 7 + shape diam. moments & 94.26 & 98.72 & 86.32 & 92.14 & 57 & 402.5 \\ 
 &16& exp 7 + shape context & 94.71 & 99.03 & 89.91 & 96.50 & 69 & 5.5 \\ 
 &17& exp 7 + convex hull & 93.47 & 98.56 & 89.97 & 96.74 & 57 & 8.7 \\ 
 &18& exp 7 + level sets overlap & 92.23 & 97.90 & 89.16 & 96.08 & 55 & 464.0 \\ 
 &19& exp 7 + level sets gradient v.f. & 93.20 & 98.28 & 89.59 & 96.17 & 53 & 35.0 \\ 
 &20& exp 7 + level sets orientation & 93.74 & 98.61 & 90.13 & 96.75 & 55 & 229.4 \\ 
 \midrule
 \multirow{4}{*}{\rotatebox{90}{Image}}
 &21& exp 7 + SIFT soft v.q.& 99.04 & 99.93 & 88.75 & 95.67 & 149 & 56.0 \\ 
 &22& exp 7 + image moments & 93.16 & 98.29 & 89.26 & 96.12 & 53 & 4.1 \\ 
 &23& exp 7 + image deriv. stats & 95.58 & 99.22 & 89.09 & 95.61 & 85 & 5.8 \\ 
 &24& exp 7 + image stats & 96.42 & 99.43 & 88.85 & 95.73 & 94 & 5.8 \\ 
 \midrule
 &25& all hand-designed ($\bigcup$ exp 1:24) & 99.98 & 99.98 & \textbf{92.33} & \textbf{97.61} & 363& - \\ 
 \bottomrule
\end{tabular}
  \caption{\small Classification experiments with hand-designed features. For each experiment, we provide the training and test accuracy (ACC), area under the ROC curve (AUC), number of total dimensions in the input
feature vector, and relative computation time for each individual feature. 
The notation `exp $X$ +' denotes that the feature set from experiment $X$ was added (i.e., set union) to the feature set in that experiment. All experiments
except 8 used a drop-out multilayer perceptron as the classifier. }
  \label{tab:hand_selected_results}

\end{table*}

\textbf{Convex hull:} the number of pixels in the convex hull of each
segment contained inside and outside of the segment and the log values of these
quantities \cite{Bas2012}.

\subsection{Classification Experiments and Results}
\label{sec:hand_designed_results}

\begin{figure}[t]
  \begin{center}
	\begin{tabular}{cc}
	\includegraphics[width=0.45\textwidth]{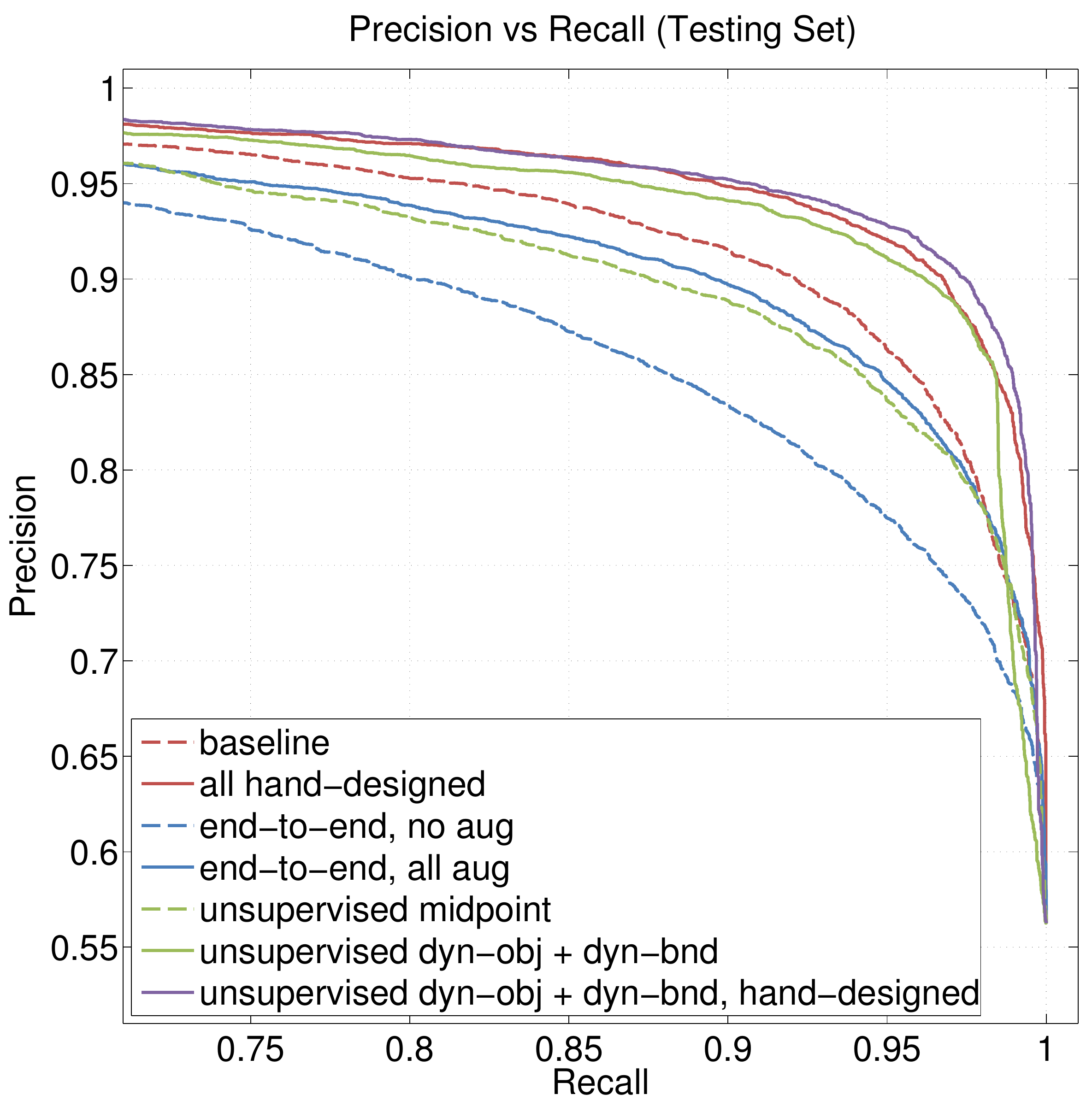} & 
	\includegraphics[width=0.45\textwidth]{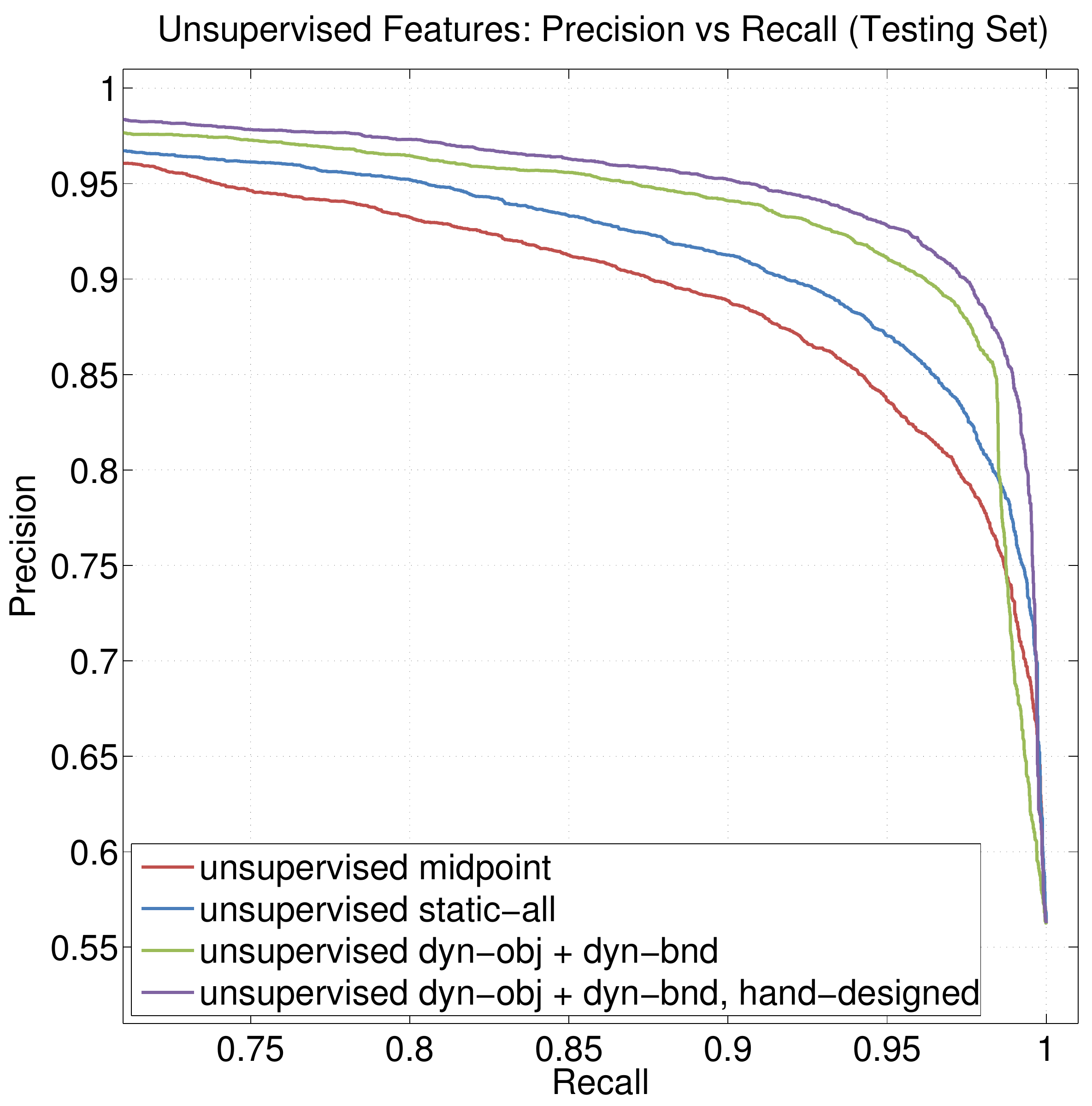} \\ 
	\small (a) Hand-Designed and Learned Experiments & \small (b) Unsupervised Learning Experiments 
\end{tabular}
  \end{center}
  \caption{
  \small Precision-recall curves comparing , and (a) hand-designed, end-to-end, and unsupervised feature learning schemes (b) different pooling schemes for unsupervised features. Unsupervised representation learning combined with dynamic pooling (\emph{unsupervised dyn--obj dyn--bnd}) yields comparable performance to an ensemble of all hand-designed features (\emph{all hand--designed}), while combining both learned and hand-designed features yields the best performance (\emph{unsupervised dyn--obj dyn--bnd, hand-designed}).  } 
  \label{fig:handselPrTraining}
\end{figure}

We performed a variety of classification experiments in which we varied the set
of hand-designed features provided to the classifier, as summarized in Table
\ref{tab:hand_selected_results}. The `Cost' column represents the wall-clock time taken to compute each feature set, normalized by the time taken for the fastest feature (`growth').  Figure~\ref{fig:handselPrTraining} shows the precision-recall curve for experiments using hand-designed features as well as results from experiments using the feature learning schemes described in subsequent sections.

 As our classifier, we use a drop-out multilayer perceptron ($200$ hidden units, $500,000$ weight updates, rectified linear hidden units) \cite{hinton2012improving}, but also present results using a decision-stump boosting classifier for comparison (experiment $8$).

Substantial improvement in performance results as the feature set increases
from a simple set of $6$ features derived from boundary map values (experiment 1:
$80.92$\% test set classification accuracy) to the combined set of all
hand-designed features (experiment 25: $92.33$\% accuracy).  Interestingly, when
considered in isolation, some of the simplest features, such as size and convex
hull, provide some of the largest improvements in accuracy. However, using all the
hand-designed features together yields significantly higher accuracy and
improved precision as compared to any individual feature. 

\section{Learned Features}
\label{sec:learned_features}

In this section, we describe two data-driven feature representations.
In contrast to hand-designed features, these representations do not
require domain knowledge specialized to the data set being considered,
and can therefore be easily adapted to new types of data.  In
addition, they are tuned to the statistics of the particular problem
being solved, and may therefore prove to be complementary to or exceed
the performance of hand-designed features.

\subsection{End-to-end Learning}
A naive but powerful approach is to simply provide the raw input
signal values to the classifier.  In such an approach, the classifier
generally consists of multiple non-linear processing layers, and the
classifier is tasked with mapping the raw input signal to intermediate
hidden representations that improve overall classification
performance.  This approach, sometimes called `end-to-end' learning,
has achieved state-of-the-art performance on a variety of vision
problems using multi-layer perceptrons and convolutional neural
networks~\cite{cirecsan2010deep, LeCun:1998}.

We implement the end-to-end learning approach in the context of 3d
agglomeration by creating, for each edge, a feature vector that
contains image, segment, and boundary information within a 3d bounding box
centered around the `decision point' (as defined at the beginning of
Section~\ref{sec:hand_designed_features}).  Specifically, we provide
raw image values from the electron microscopy data, boundary map
values, and two binary segment masks. A particular mask is non-zero
only where a given segment belonging to the edge is present.  

A multiscale representation of the region around the decision point
can be obtained by extracting the raw voxel values using multiple
windows of varying radii.  Further, to control the dimensionality of
the input when using a large window radius, the subvolume of raw
values can be downsampled by some factor $d$ in each spatial
dimension.  As a result, for a particular scale consisting of a
bounding box of radius $r$ and downsampling of $d$, the total
dimensionality of the feature vector is $4 \times (2\frac{r}{d}+1)^3$.

\subsection{Unsupervised Feature Learning}

End-to-end learning can be particularly difficult when the size of the
training set is limited (relative to the dimensionality of the data),
as the classifier must discover useful patterns and invariances in the
original data representation from a limited amount of supervised
signal.  However, the original (unlabeled) data itself can be useful
as an additional signal, by learning representations that are capable
of reproducing the data.  These `unsupervised' approaches learn
feature representations by optimizing models that reconstruct the raw
data in the presence of various forms of regularization such as a
bottleneck or sparsity~\cite{Hinton:2006, olshausen1996emergence}.

\begin{figure}[t]
  \begin{center}
    \begin{tabular}{cccc}
		\raisebox{.25\height}{\rotatebox{90}{\parbox{0.7in}{\centering Dynamic\\Object\\pooling}}} &
      \raisebox{.25\height}{\frame{\includegraphics[height=0.7in]{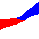}}} &
      \raisebox{.25\height}{\frame{\includegraphics[height=0.7in]{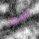}}} &
      \includegraphics[height=1.0in]{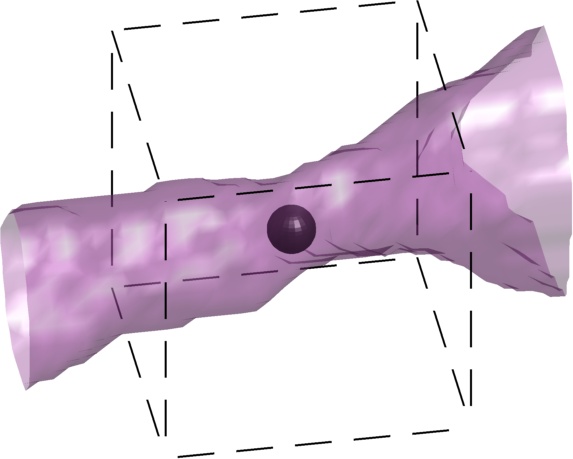} \\
		\raisebox{.25\height}{\rotatebox{90}{\parbox{0.7in}{\centering Dynamic\\Boundary\\pooling}}} &
      \raisebox{.25\height}{\frame{\includegraphics[height=0.75in]{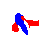}}} &
      \raisebox{.25\height}{\frame{\includegraphics[height=0.75in]{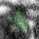}}} &
      \includegraphics[height=1.0in]{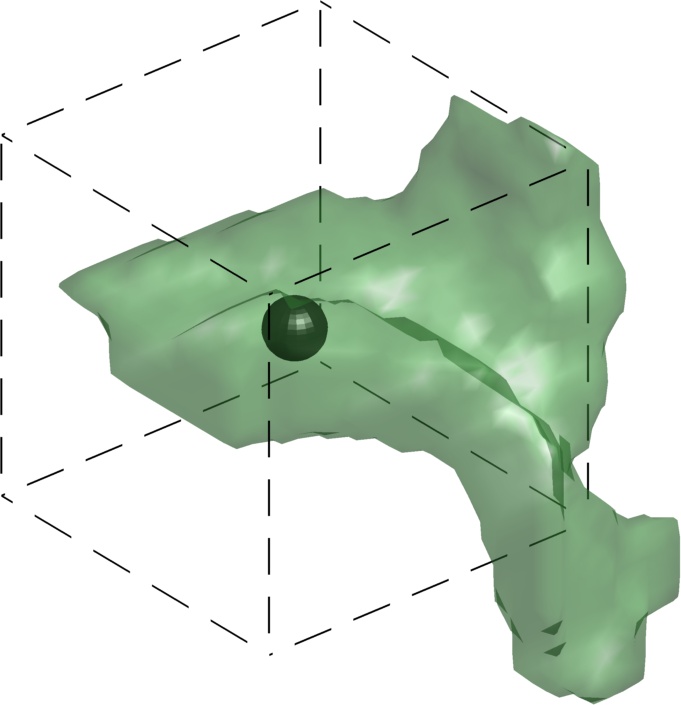} \\
		& Segments & Pooling region & Rendering \\
    \end{tabular}
  \end{center}
  \caption{\small Examples of dynamic pooling: the top row shows object
    pooling for the positive edge segments shown in
    Figure~\ref{fig:exampleRenderings}, and the bottom row shows
    boundary pooling for the negative edge segments.  The left column
    shows 2d $x$-$y$ slices of the segmentation, and the center column
    shows the corresponding raw image data with an overlay of the
    pooling region, where the dynamic pooling regions correspond to
    using a window of radius 10 voxels, and the total slice area
    corresponds to the context needed to generate the feature
    representation for all locations in the window.  The right column
    gives a rendering of the 3d pooling regions, where the pooling
    window is given by the bounding box indicated by dashed lines.}
  \label{fig:dyn_pool}
\end{figure}


We experimented with using the unsupervised feature learning and
extraction module used in DAWMR networks and adapting it to the
agglomeration task.  The core of this module consists of vector
quantization (VQ) of $5^3$ patches of the data, where the dictionary
is learned using orthogonal matching pursuit (OMP-1), and encoding is
performed using soft-thresholding with reverse polarity.  This core
component is performed at two scales (original resolution and
downsampling by two in each spatial dimension), and a foveated
representation is produced by concatenating the encoding produced at a
center location with a max-pooled encoding over all locations within a
radius of two of the center.  Therefore, a $9^3$ support region is
used to produce the representation centered at a given voxel.

A straightforward method of adapting this feature representation to
the problem of 3d agglomeration is to simply extract the feature
representation at the decision point, which we will refer to as simply
the `midpoint' feature.  Similar to end-to-end learning described
above, the input data to the feature learning and extraction module
consists of the raw image values, boundary map values, and a single
binary segment mask that is non-zero only where either segment
belonging to the edge is present.  (We found that a single binary
segment mask gave comparable performance to using two separate masks
as used in end-to-end learning.)

However, the agglomeration task of deciding whether or not to merge
two segments likely requires a greater context than the boundary prediction
problem that the DAWMR feature representation was originally designed for.
Therefore, we also consider extracting the foveated feature representation from
every location within a fixed-radius window of the decision point, and
average-pool these features.  We refer to this as `static-all' pooling, and concatenate this feature with the midpoint feature to obtain the `midpoint + all' feature set.

We further introduce the notion of \emph{dynamic pooling}, where the
region to pool over is dependent on the segments themselves.  For
instance, rather than average pooling over all features within a
window as in `all' pooling, we can restrict the average pooling to be
over only features corresponding to locations in either of the two
segments (within a fixed-radius window of the decision point).  This
procedure, which we term `dyn-obj' dynamic object pooling, may improve results
over `all' pooling by ignoring locations that are irrelevant to the
agglomeration decision.

Another approach to dynamic pooling is to focus on those locations
whose interpretation would change as a result of the agglomeration
decision.  In particular, the interpretation of those locations
`in-between' the two segments would change depending on whether the
two segments were merged into a single object or kept as two separate
segments.  Therefore, we introduce the notion of dynamic pooling along the
boundary between two segments, which we refer to as `dyn-bnd' pooling.
This is done by dilating each segment by a fixed amount (in our
experiments, by half the radius used for the window around the
decision point), and then considering those locations in the
intersection of the two dilated segments.  Both dynamic pooling
methods are illustrated in Figure~\ref{fig:dyn_pool}.

Finally, similar to end-to-end learning above, we consider multiscale
dynamic pooling representations given by extracting features within
windows of differing radii.

\subsection{Classification Experiments and Results}
\label{sec:learned_results}

\begin{table*}[t]

\setlength\tabcolsep{3pt}
\centering
\begin{tabular}{ c r l c c c c r r }
\toprule
 & & & \multicolumn{2}{c}{Training set} & \multicolumn{2}{c}{Testing set} & \\
\cmidrule(r){4-5}\cmidrule(r){6-7}
 &Exp.& Feature Set Description & ACC(\%) & AUC(\%) & ACC(\%) & AUC(\%) & Dim. & Cost\\ 
\midrule 
\multirow{5}{*}{ \rotatebox{90}{end-to-end}}
 &1& $r$=5, $d$=1, 20 hidden units  & 98.87 & 99.64 & 82.62 & 92.36 &5324&7.6\\
 &2& $r$=5, $d$=1  						& 100.0 & 99.98 & 84.34 & 93.53 &5324&7.6\\
 &3& $r$=5, $d$=1, 400 hidden units & 100.0 & 99.98 & 85.02 & 93.62 &5324&7.6\\
 &4& exp 2 + ($r$=10, $d$=2) 			& 100.0 & 99.98 & 84.50 & 93.75 &10,648&12.6\\
 &5& exp 3 + ($r$=10, $d$=2) 			& 100.0 & 99.98 & 85.54 & 93.99 &10,648&12.6\\
\midrule
\multirow{5}{*}{ \rotatebox{90}{unsupervised}}
 &6&  midpoint                               & 100.0 & 99.98 & 87.84 & 95.19 &  8000 &   9.9\\ 
 &7&  midpoint + static-all ($r$=10)                & 100.0 & 99.98 & 88.85 & 95.89 & 16,000 & 371.7\\ 
 &8&  midpoint + dyn-obj ($r$=10)                & 100.0 & 99.98 & 89.65 & 96.28 & 16,000 & 368.0\\ 
 &9&  dyn-bnd ($r$=10)                           & 100.0 & 99.98 & 91.24 & 96.96 &  8000 & 371.2 \\ 
 &10& dyn-obj + dyn-bnd ($r$=4 + $r$=10)& 100.0 & 99.98 & \textbf{91.38} & \textbf{97.14} & 16,000 & 246.6\\ 
\bottomrule
\end{tabular}
  \caption{\small Classification experiments with learned features. Dynamic pooling strategies (\emph{dyn-obj} and \emph{dyn-bnd}) are critical to achieving accuracy levels competitive with hand-designed features. }

 \label{tab:learned_feature_results}

\end{table*}

Results using the two data-driven representations above are presented in
Table~\ref{tab:learned_feature_results} and Figure~\ref{fig:handselPrTraining}(a).  As with the hand-designed features,
classification was performed using a drop-out multilayer perceptron, with $200$
hidden units unless otherwise specified. The end-to-end features outperform
rudimentary boundary map features for the test set but not the larger feature
set. The unsupervised feature set achieves much better test set performance,
approaching that of the all hand-designed features.  Figure~\ref{fig:handselPrTraining}(b)
demonstrates the improvements that dynamic pooling methods can achieve.

\section{Training Set Augmentation}
\label{sec:augmentation}

Next, we describe experiments designed to improve generalization
performance through synthetic augmentation of the training set.  The
motivation behind this methodology comes from work such
as Decoste and Sch\"{o}lkopf~\cite{Decoste2002} and Drucker~\etal~\cite{Drucker1993}, which create `virtual'
examples by applying some set of transformations to examples in the
original training set and use these examples during classifier
training.  Thus, the training procedure is more likely to produce a
classifier invariant to the given transformations. In this work, we experiment with `swap,' `isometry,' and `jitter' augmentations of the training data.  

The `swap' transformations exchange the identities of the first and second
segments (\ie swapping the ordering). The `isometry' augmentation considers all possible isometries of the underlying data.  The image data is slightly
anisotropic, as the $z$-axis corresponds to the milling direction in
FIB-SEM, orthogonal to the imaging plane (see Section~\ref{sec:EM},
Electron microscopy images).  Distance-preserving maps of the data
therefore include four $90^\circ$ rotations of the $x$-$y$ plane,
reflection of the $x$-axis, and reflection of the $z$-axis.  In total,
these transforms form a group of order 16, equivalent to the
isometries of a square prism, or $\text{D}_{4\text{h}}$. Finally, `jitter' augmentations slightly shift the location of the decision
point.  In this work, our experiments use $27$ different decision points, where
the original decision point is offset by all combinations of $\lbrace -1, 0, 1
\rbrace$ in all coordinates.

\subsection{Augmentation Results}
\label{sec:aug_results}

We experimented with hand-designed, end-to-end, and unsupervised features  using training set augmentation. We also explored using a more powerful classifier with two hidden layers; this deeper classifier could be especially important when augmentation is used, as the amount of training data increases dramatically. 

Table~\ref{tab:augmented_results} in the supplementary gives the full results for experiments with different types of augmentation.  Although augmented training examples had a slightly detrimental effect on the classification results when using hand-designed features, the end-to-end features benefited significantly by using all augmentation types simultaneously (thus expanding training set size from $14,552$ to $12,572,928$ examples). The two-layer MLP classifier further improved performance for end-to-end features using all augmentations. Overall, however, even after including augmentations in the end-to-end experiments, generalization performance was still much worse as compared to unsupervised feature experiments performed \emph{without} augmentation. The unsupervised feature experiments that included augmentations saw minimal effects on generalization performance.

\section{Feature Combination and Selection}

We explore combining various learned feature schemes with hand-designed
features, with the hypothesis that the hand-designed features may more easily
capture higher-level or non-linear edge or segment characteristics than the learned methods.
We use all training set augmentations for these experiments, since this case
markedly improved the end-to-end feature learning approach. For computational
reasons, we omitted the most expensive hand-designed features, namely, SIFT
features, shape diameter (moments, and quantiles), level set overlap, and level
set orientation. Proximity was not omitted because it is necessary for other
aspects of the pipeline. Results of these
experiments are given in Table~\ref{tab:combination_results}. Test set accuracy
improves for both end-to-end and unsupervised learned features when used in
combination with hand-designed features, though the improvement is more marked
for end-to-end features.

\begin{table*}[t] 
\centering
\setlength\tabcolsep{2pt}
\begin{tabular}{ c l c c c c r r }
\toprule
 & & \multicolumn{2}{c}{Training set} & \multicolumn{2}{c}{Testing set} & & \\
 \cmidrule(r){3-4}\cmidrule(r){5-6}
 Exp. & Feature Set Description & ACC(\%) & AUC(\%) & ACC(\%) & AUC(\%) &
 Dim. & Training Ex. \\
\midrule
 1 & hand-designed ~+~end-to-end  & 95.06 & 99.07 & 92.09 & 97.74 & 5546 & 12,572,928 \\ 
 2 & hand-designed ~+~unsupervised  & 100.0 & 99.98 & 92.21 & 97.67 & 16,222 & 1,571,616 \\ 
\bottomrule
\end{tabular}
  \caption{ \small 
  Classification experiments using a combination of hand-designed and learned features.}

  \label{tab:combination_results}
\end{table*}

\section{Discussion}

We have demonstrated that features derived purely from learning algorithms can provide highly informative representations for a classification task involving 3d objects. The key innovation in achieving this result was a type of dynamic pooling that selectively pools feature representations from different spatial locations in a manner (dynamically) dependent on the shape of the underlying objects involved in the classification. We were able to implement this strategy in a straightforward way using an unsupervised learning approach, as the feature learning phase was separated from the encoding stage in which the pooling is performed.

These methods and results are a starting point for further work involving feature learning methods applied to 3d objects. In particular, the results motivate a more sophisticated end-to-end strategy that also incorporates dynamic pooling. Learning such an architecture will be more involved than in the unsupervised case, as the variations in spatial pooling (from one example to the next) will need to be incorporated into the learning algorithm. 

Another open question is whether learning architectures for these types of problems would benefit from more complicated non-linearities or recurrent/recursive structure; some of the hand-designed features that appear to provide predictive benefit are based on highly non-linear iterative methods (e.g, level sets) or ray-tracing (e.g., ray features and shape diameter function), both of which are computations that might be difficult for a typical multilayer network architecture to emulate. Adding specific representational capacity motivated by these hand-designed strategies while preserving the ability to train most details of the architecture could offer a superior approach.

{\small
\textbf{Acknowledgements:} We thank Zhiyuan Lu for sample preparation, Shan Xu and Harald Hess for FIB-SEM imaging, and Corey Fisher and Chris Ordish for data annotation.
}

{\small{\bibliographystyle{ieee}
\bibliography{bib/curr_opinion,bib/Features,bib/EM,bib/Agglomeration,bib/Software,bib/Trees,bib/DeformableModels,bib/SVM,bib/MLP_ANN}
}}

\newpage
\appendix
\section{Supplementary: Edge examples}
\begin{figure}[h]
  \begin{center}
   \begin{tabular}{ccc}
	\raisebox{.5\height}{\rotatebox{90}{\parbox{0.7in}{\centering Positive\\Edge }}} &
	\includegraphics[height=1.7in,angle=90]{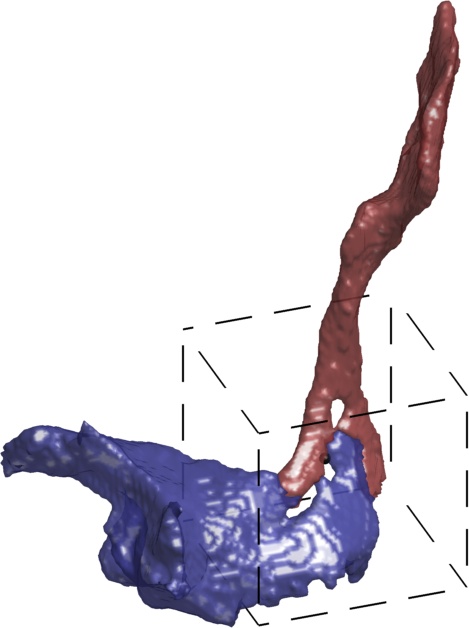} & 
	\includegraphics[width=1.85in]{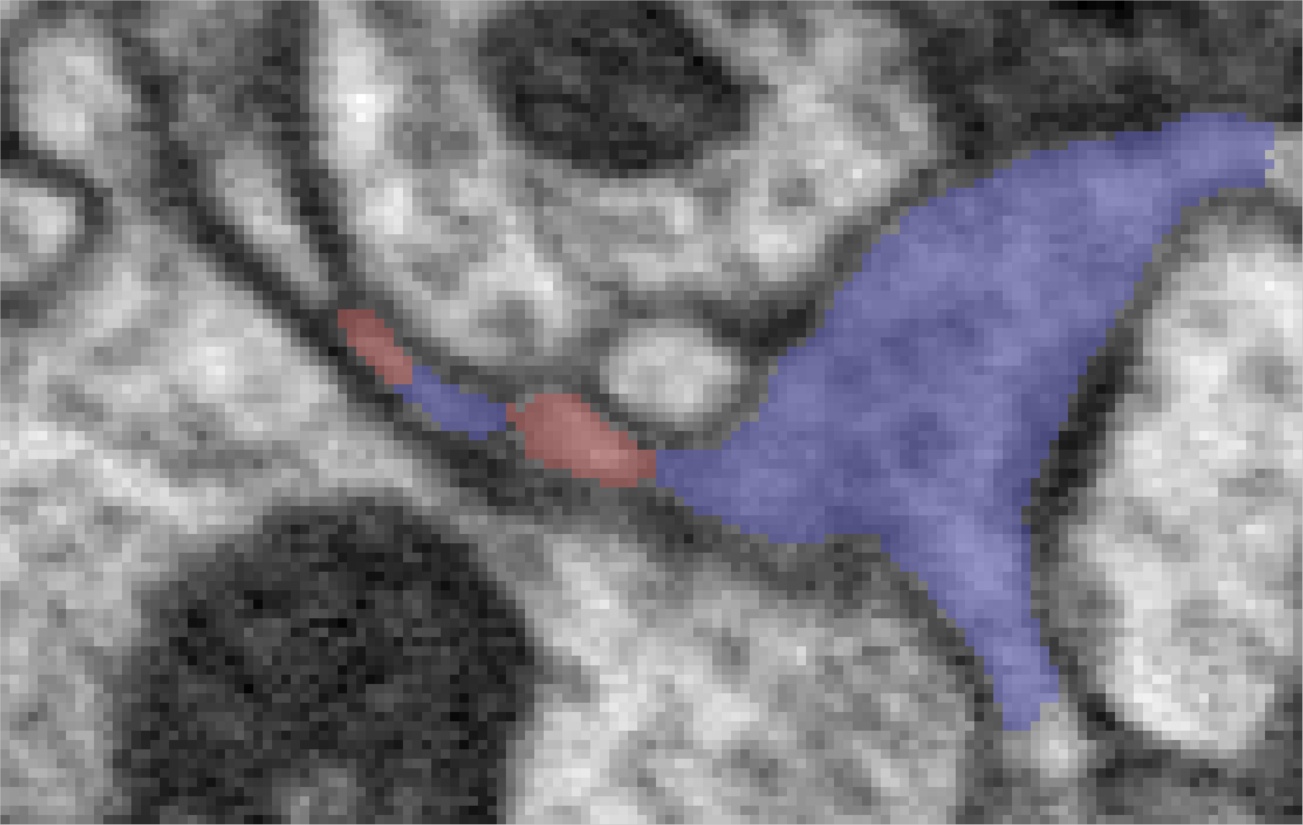}\\  
	\raisebox{.75\height}{\rotatebox{90}{\parbox{0.7in}{\centering Negative\\Edge }}} &
	\includegraphics[height=1.7in]{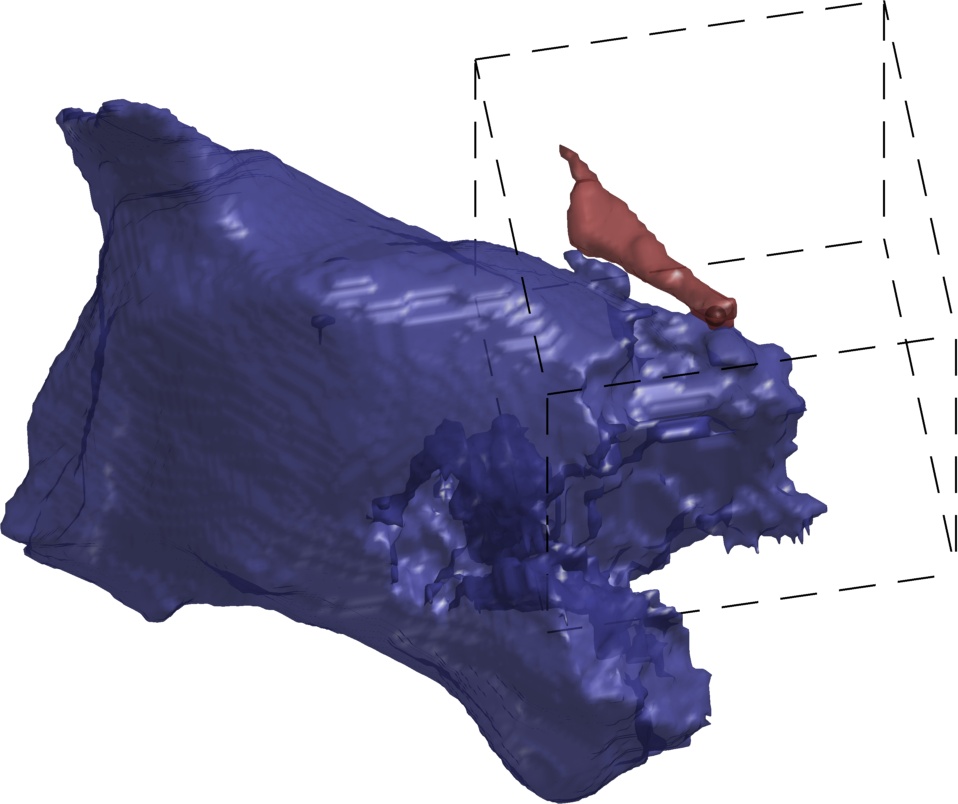} & 
	\includegraphics[height=1.85in,angle=90]{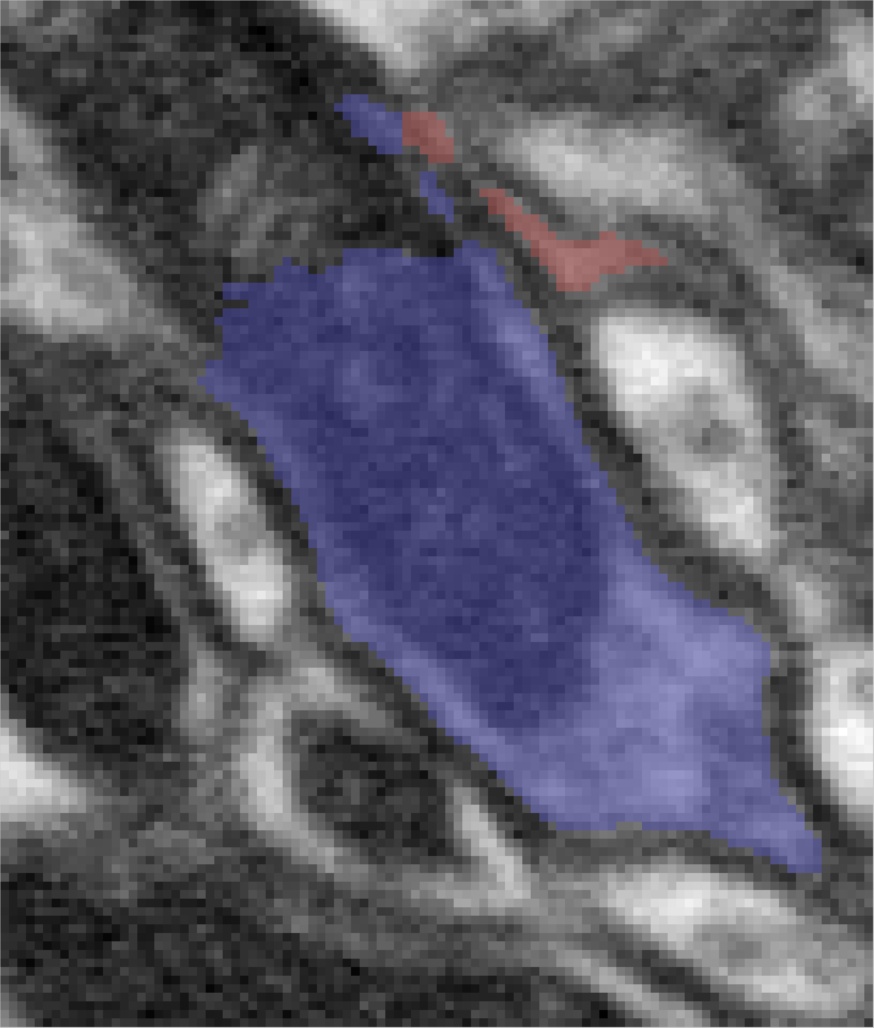} \\ 
	& Rendering & Slice \\
  \end{tabular}
  \end{center}
  \caption{ 
	 Complex 3d segment shapes and spatial relationships makes agglomeration more challenging.
	 The interdigitation of the segments in the positive example create a complex
	 interface. The negative edge demonstrates that very thin structures can abut
	 large segments, creating a interface relative to the smaller object.
	 Furthermore, since much of the larger segment lies outside of the decision
    window, global properties (e.g., orientation) cannot be computed.
     }
  \label{fig:extra_edge_examples}
\end{figure}

\section{Supplementary: Error Analysis}
\label{sec:error_hs}

In this section, we explore potential causes of classifier errors and make 
suggestions for future improvements of the pipeline.  Specifically, we examine
a set of $50$ edges, half of which were false positives, half of which were
false negatives, for which the both the MLP classifier and human expert were
confident. We manually examined the segments, image, and affinity graph for
these edges in an attempt to glean potential patterns that might help drive
future improvements. 

\begin{figure}[h]
  \begin{center}
   \begin{tabular}{cc}
		\includegraphics[height=1.8in]{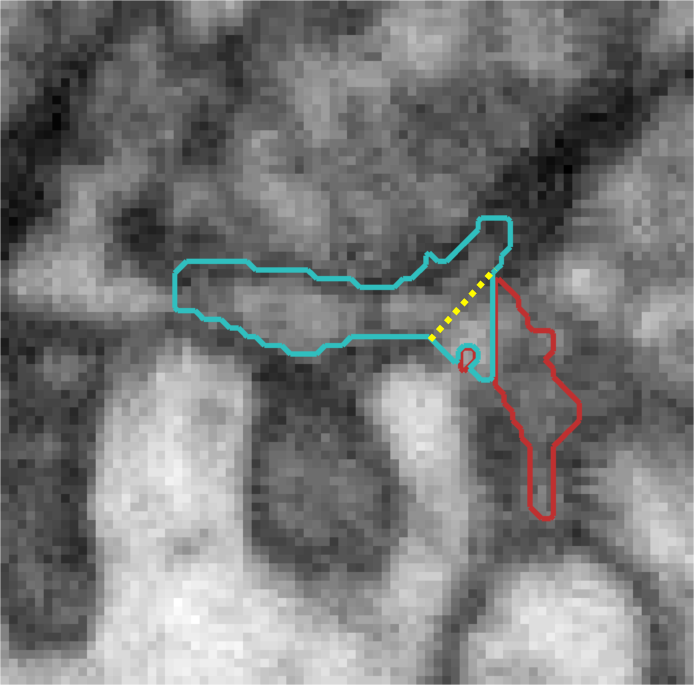} &
		\includegraphics[height=1.8in]{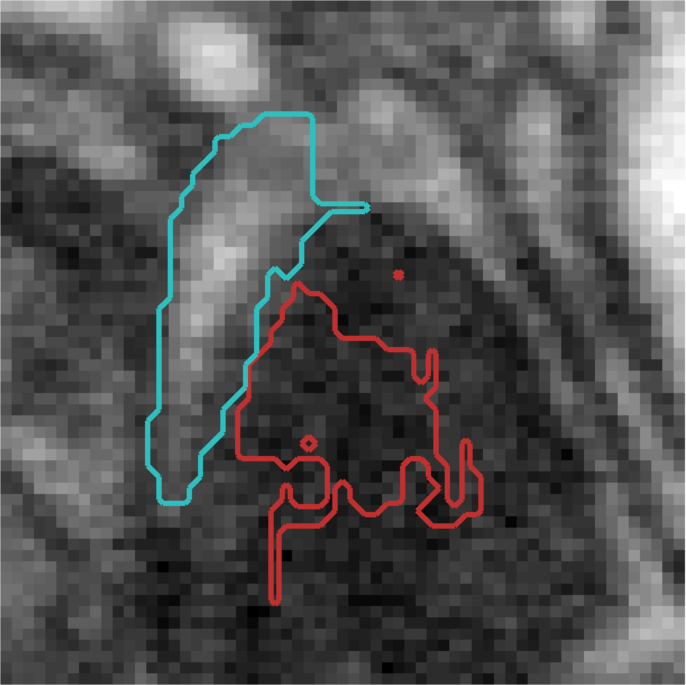} \\ 
        (a) False positive & (b) False negative \\
   \end{tabular}
  \end{center}
  \caption{Examples of false positive and false negative edges for experiment 25
	 in Table~\ref{tab:hand_selected_results}. The yellow line in (a) shows the
	 true boundary between cells that have been undersegmented.}
  \label{fig:error_hs_examples}
\end{figure}

One characteristic that seems common among errors is the presence of
`undersegmentation,' the presence of segments that overlap more than one true object. 
Undersegmentation appear in $15$ of the $50$
error cases; $13$ of those examples are false positives.  It is
possible that these errors are due to segments that erroneously grow
across cell boundaries. This can cause segments to become adjacent
when the true objects are not, thereby confusing the classifier.  An example of
this phenomenon is shown in Figure~\ref{fig:error_hs_examples}(a).

Another property of some errors seems to be that they occur near boundaries of
internal cell structures, such as mitochondria.
Figure~\ref{fig:error_hs_examples}(b) shows an example of such a false negative
edge.  Notice that in this example, the red segment lies inside a mitochondrion, 
the blue segment consists of part of the cell outside the mitochondrion, and the
two segments share a mutual boundary.

The patterns of error we observed above suggest some improvements for future work.
First, the undersegmentations could be ameliorated either by refinements to the boundary
prediction or to the procedure that generates segments from the boundaries. 
Segments that are too large could cause some of the false positives we have observed, 
and suggests that using a more conservative oversegmentation scheme that yields
smaller objects might be preferable. Of course, whether this approach would cause
false negatives would need exploration.

Secondly, the errors occurring near mitochondrial and other intra-cellular boundaries
suggest that our methodology might benefit from a framework that explicitly identifies
the locations of these problem areas.  This new information could improve agglomeration, 
boundary prediction, or both.

\section{Supplementary: Precision-Recall Plots}
\label{sec:supp_PR}
\begin{figure}[h]
  \begin{center}
   \begin{tabular}{cc}
	\includegraphics[width=0.45\textwidth]{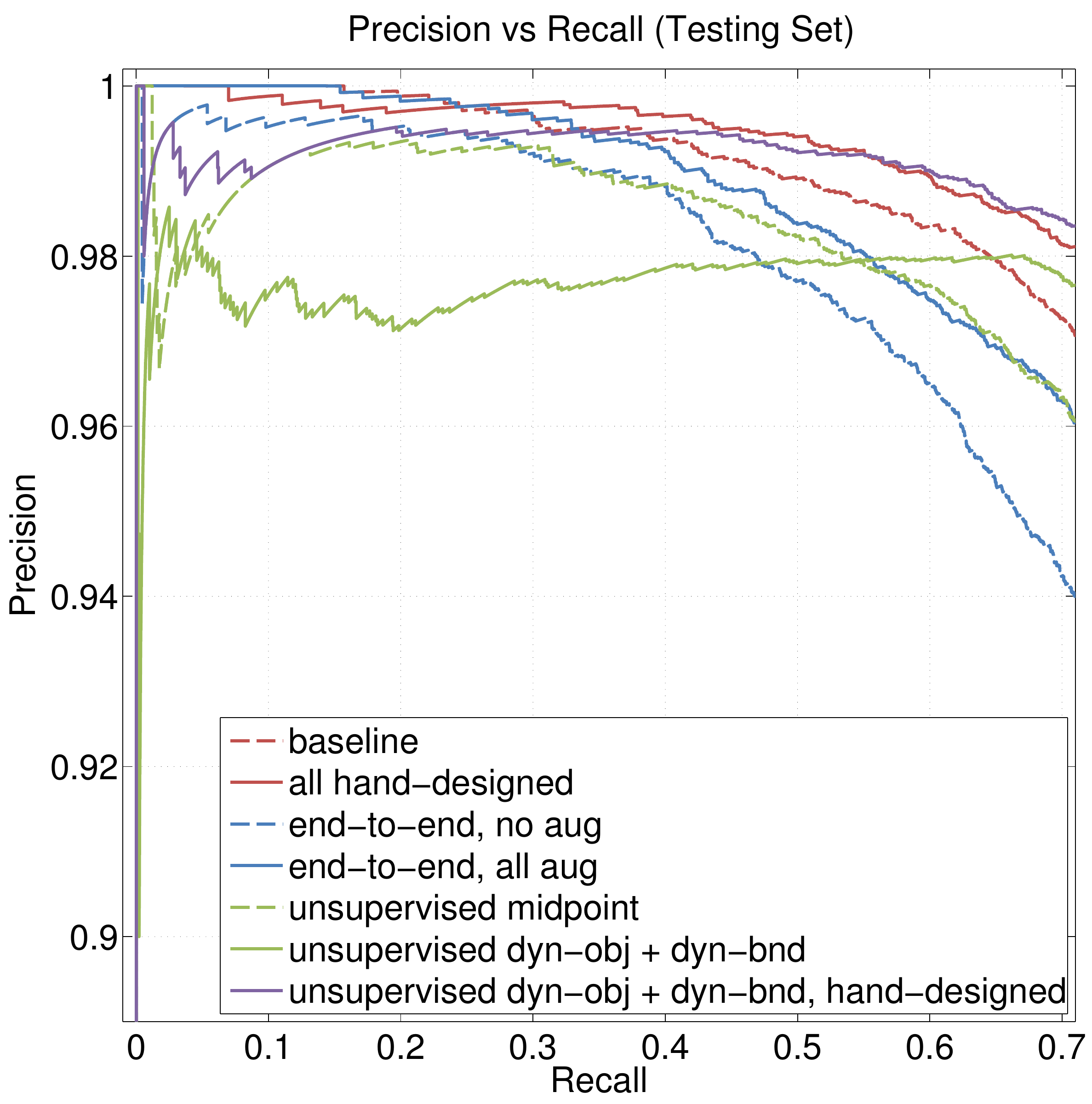} & 
	\includegraphics[width=0.45\textwidth]{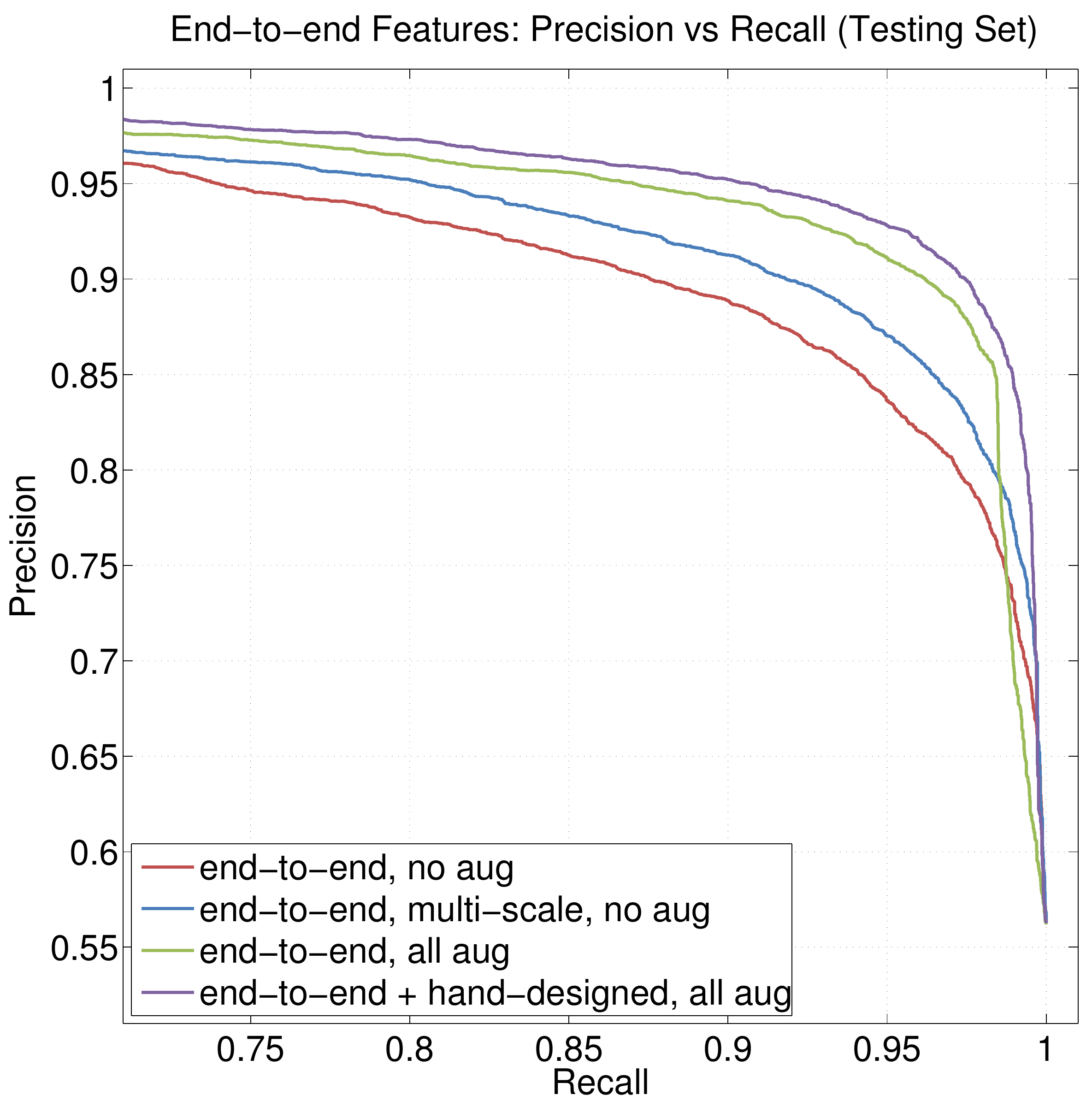} \\ 
   (a) Low-recall region & (b) End-to-end  \\  
  \end{tabular}
  \end{center}
  \caption{ 
	 Precision-recall curves in (a) low-recall regions and (b) for the
    end-to-end learned feature scheme. }
  \label{fig:PR_supp}
\end{figure}

Figure~\ref{fig:PR_supp}(a) shows the precision-recall curves in the low recall 
region.  Experiments with very high training-set accuracy do not achieve high 
precision on the testing set due to overfitting.  Figure~\ref{fig:PR_supp}(b)
shows the precision-recall curves for end-to-end feature learning.  
Including training set augmentation improves performance
much more than a multi-scale approach.  Combining end-to-end features with
hand-designed features improved the end-to-end features significantly. 

\pagebreak
\section{Supplementary: Full Augmentation Results}
\begin{table*}[h] 
\setlength\tabcolsep{2pt}
\centering
\begin{tabular}{ c r l c c c c r r }
\toprule
 & & & \multicolumn{2}{c}{Training set} & \multicolumn{2}{c}{Testing set} & & \\
 \cmidrule(r){4-5}\cmidrule(r){6-7}
 Aug. & Exp. & Feature Set Description & ACC(\%) & AUC(\%) & ACC(\%) & AUC(\%) &
 Dim. & Training Ex. \\
\midrule
\multirow{3}{*}{ \rotatebox{90}{None}}
 & 1 & All Hand-sel. & 99.98 & 99.98 & 91.04 & 96.33 &   363 & 14,552\\
 & 2 & End-end       & 100.0 & 99.98 & 84.34 & 93.53 &  5324 & 14,552\\ 
 & 3 & Unsup.~Learned   & 100.0 & 99.98 & 91.38 & 97.14 & 16,000 & 14,552\\
\midrule
\multirow{3}{*}{ \rotatebox{90}{Swap}}
 & 4 & All Hand-sel. & 100.0 & 99.98 & 90.91 & 96.39 & 363 &  29,104\\ 
 & 5 & End-end 		& 100.0 & 99.98 & 84.93 & 93.07 & 5324 &  29,104 \\ 
 & 6 & Unsup.~Learned 	& \multicolumn{6}{c}{invariant to segment order} \\
\midrule
\multirow{3}{*}[1.5pt]{\rotatebox{90}{Isometry}}
 & 7 & All Hand-sel. & 99.98 & 99.98 & 91.13 & 96.49 & 363   & 232,832\\ 
 & 8 & End-end     	& 97.91 & 99.77 & 84.85 & 93.30 & 5324  & 232,832\\ 
 & 9 & Unsup.~Learned   & 99.84 & 99.98 & 91.36 & 97.27 & 16,000 & 232,832\\
\midrule
\multirow{3}{*}[-1pt]{ \rotatebox{90}{Jitter}}
 & 10 & All Hand-sel. & 100.0 & 99.98 & 90.06 & 95.18 & 363   &  392,904\\ 
 & 11 & End-end       & 99.96 & 99.98 & 85.37 & 93.51 & 5324  & 392,904\\ 
 & 12 & Unsup.~Learned   & 100.0 & 99.98 & 91.35 & 97.18 & 16,000 &  392,904\\
\midrule
\multirow{2}{*}[-1pt]{ \rotatebox{90}{All}}
 & 13 & End-end 			& 89.48 & 96.32 & 86.38 & 94.55 &   5324 & 12,572,928\\ 
 & 14 & Unsup.~Learned 	& 100.0 & 99.98 & 91.57 & 97.41 & 16,000 &  1,571,616\\ 
\midrule
\multirow{3}{*}{ \rotatebox{90}{None }}
 & 15 & All Hand-sel.~(MLP2) 	& 100.0 & 99.98 & 91.07 & 96.01 &   363 & 14,552\\ 
 & 16 & End-end (MLP2)			& 100.0 & 99.98 & 84.48 & 92.67 &  5324 & 14,552\\ 
 & 17 & Unsup.~Learned (MLP2) & 100.0 & 99.98 & 91.77 & 97.13 & 16,000 & 14,552\\
\midrule
\multirow{2}{*}[-1pt]{ \rotatebox{90}{All}}
 & 18 & End-end        (MLP2) & 91.99 & 97.54 & 88.00 & 95.34 & 5324 & 12,572,928\\ 
 & 19 & Unsup.~Learned (MLP2) & 99.98 & 99.98 & 91.50 & 97.24 & 16,000 & 1,571,616\\ 
\bottomrule
\end{tabular}
  \caption{ \small Classification experiments using augmented training data and MLP's
  with two hidden layers. The hand-designed feature set is comparable to
experiment $25$ in Table~\ref{tab:hand_selected_results}, end-to-end features are  are comparable to experiment $2$ in Table~\ref{tab:learned_feature_results}, and unsupervised features are comparable to experiment $10$ in Table~\ref{tab:learned_feature_results}. For computational reasons, we omit the `all augmentations' experiment using all hand-designed features.}

  \label{tab:augmented_results}

\end{table*}

\end{document}